\documentclass[letterpaper, 10 pt, conference]{ieeeconf}  %

\IEEEoverridecommandlockouts                              %

\title{\LARGE \bf
Composable Deep Reinforcement Learning for Robotic Manipulation
}

\author{Tuomas Haarnoja$^1$, Vitchyr Pong$^1$, Aurick Zhou$^1$, Murtaza Dalal$^1$, Pieter Abbeel$^{1,2}$, Sergey Levine$^{1}$ %
\thanks{$^{1}$Berkeley Artificial Intelligence Research, UC Berkeley, $^{2}$Open AI
        {\newline\tt\small \{haarnoja, vitchyr, azhou42, mdalal, pabbeel, svlevine\}@berkeley.edu}}%
	}

\usepackage{amsfonts}
\usepackage{amsopn}
\usepackage{amsmath}
\usepackage{amssymb}
\usepackage{mathtools}
\usepackage{bbm}
\usepackage{color}
\usepackage{nicefrac}
\usepackage{placeins}

\makeatletter  
\let\NAT@parse\undefined
\makeatother

\usepackage{hyperref}

\newtheorem{theorem}{Theorem}
\newtheorem{lemma}{Lemma}
\newtheorem{corollary}{Corollary}

\newcommand{\eg}{e.g.\ }

\newcommand{\etal}{et al.\ }

\newcommand{\E}[2]{\operatorname{\mathbb{E}}_{#1}\left[#2\right]}

\newcommand{\Dalpha}[2]{\mathcal{D}_{\frac{1}{2}}\left(#1\;\middle\|\;#2\right)}
\newcommand{\softmax}[1]{\underset{#1}{\operatorname{\,softmax\,}}}

\newcommand{\ent}{\mathcal{H}}
\newcommand{\sdots}{\,\cdot\,}
\newcommand{\voidarg}{\,\cdot\,}

\newcommand{\sspace}{\mathcal{S}}
\newcommand{\aspace}{\mathcal{A}}
\newcommand{\state}{\mathbf{s}}

\newcommand{\st}{{\state_t}}

\newcommand{\stp}{{\state_{t+1}}}

\newcommand{\action}{\mathbf{a}}

\newcommand{\at}{{\action_t}}

\newcommand{\opt}{^*}

\newcommand{\reward}{r}

\newcommand{\Q}{Q}

\newcommand{\policy}{\pi}

\newcommand{\discount}{\gamma}

\newcommand{\arefA}[1]{\hyperref[#1]{Appendix~A}}
\newcommand{\arefB}[1]{\hyperref[#1]{Appendix~B}}
\newcommand{\aref}[1]{\hyperref[#1]{Appendix~\ref{#1}}}

\def\alignautorefname~#1\null{%
  (#1)\null
}\def\equationautorefname~#1\null{%
  Equation~#1\null
}

\begin{document}

\maketitle
\thispagestyle{empty}
\pagestyle{empty}

\begin{abstract}
Model-free deep reinforcement learning has been shown to exhibit good performance in domains ranging from video games to simulated robotic manipulation and locomotion. However, model-free methods are known to perform poorly when the interaction time with the environment is limited, as is the case for most real-world robotic tasks. In this paper, we study how maximum entropy policies trained using soft Q-learning can be applied to real-world robotic manipulation. The application of this method to real-world manipulation is facilitated by two important features of soft Q-learning. First, soft Q-learning can learn multimodal exploration strategies by learning policies represented by expressive energy-based models. Second, we show that policies learned with soft Q-learning can be composed to create new policies, and that the optimality of the resulting policy can be bounded in terms of the divergence between the composed policies. This compositionality provides an especially valuable tool for real-world manipulation, where constructing new policies by composing existing skills can provide a large gain in efficiency over training from scratch. Our experimental evaluation demonstrates that soft Q-learning is substantially more sample efficient than prior model-free deep reinforcement learning methods, and that compositionality can be performed for both simulated and real-world tasks.
\end{abstract}

\section{INTRODUCTION}
The intersection of expressive, general-purpose function approximators, such as neural networks, with general purpose model-free reinforcement learning algorithms that can be used to acquire complex behavioral strategies holds the promise of automating a wide range of robotic behaviors: reinforcement learning provides the formalism for reasoning about sequential decision making, while large neural networks provide the representation that can, in principle, be used to represent any behavior with minimal manual engineering. However, applying model-free reinforcement learning algorithms with multilayer neural network representations (i.e., deep reinforcement learning) to real-world robotic control problems has proven to be very difficult in practice: the sample complexity of model-free methods tends to be quite high, and is increased further by the inclusion of high-capacity function approximators. Prior work has sought to alleviate these issues by parallelizing learning across multiple robots~\cite{gu2017deeprobot}, making use of example demonstrations~\cite{finn2016connection,vevcerik2017leveraging}, or training in simulation and relying on an accurate model to enable transfer to the real world~\cite{rusu2016progressive,james2017transferring}.
All of these approaches carry additional assumptions and limitations. Can we instead devise model-free reinforcement learning algorithms that are efficient enough to train multilayer neural network models directly in the real world, without reliance on simulation, demonstrations, or multiple robots?

\begin{figure}[t] 
  \includegraphics[height=45mm]{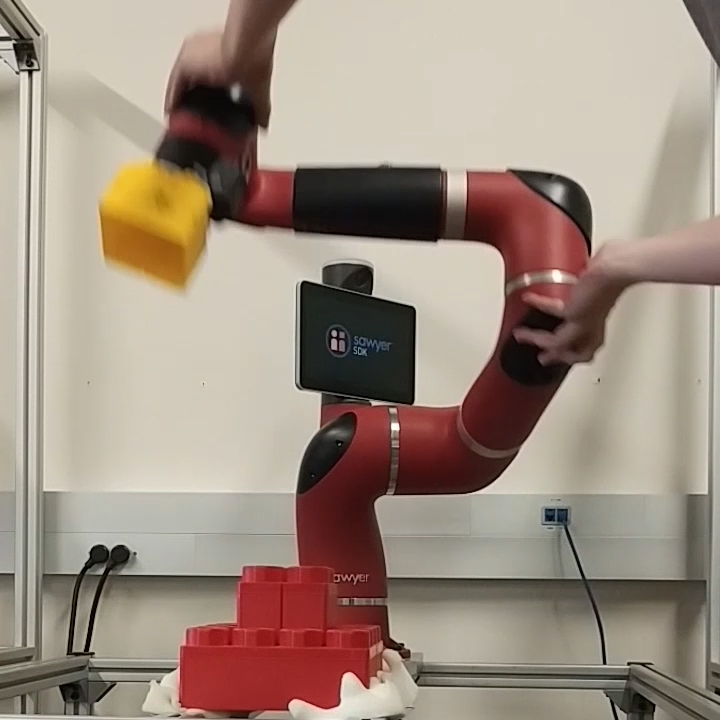}\hspace{2mm} %
  \includegraphics[height=45mm,trim={0 5mm 0 50mm},clip]{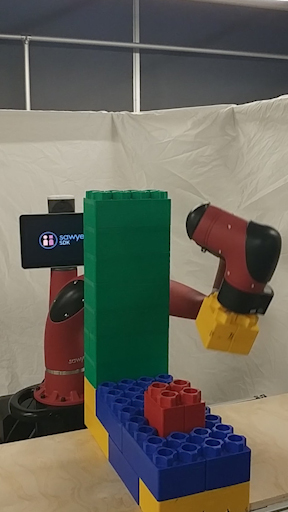}
\caption{
We trained a Sawyer robot to stack Lego blocks together using maximum entropy reinforcement learning algorithm called soft Q-learning. Training a policy from scratch takes less than two hours, and the learned policy is extremely robust against perturbations (left figure). We also demonstrate how learned policies can be combined to form new compound skills, such as stacking while avoiding a tower of Lego blocks (right figure).}
\label{fig:sawyer}
\vspace{-0.6cm}
\end{figure}

We hypothesize that the maximum entropy principle~\cite{ziebart2008maximum} can yield an effective framework for practical, real-world deep reinforcement learning due to the following two properties. First, maximum entropy policies provide an inherent, informed exploration strategy by expressing a stochastic policy via the Boltzmann distribution, with the energy corresponding to the reward-to-go or Q-function~\cite{haarnoja2017reinforcement}. This distribution assigns a non-zero probability to all actions, but actions with higher expected rewards are more likely to be sampled. As a consequence, the policy will automatically direct exploration into regions of higher expected return. This property, which can be thought of as a soft combination of exploration and exploitation, can be highly beneficial in real-world applications, since it provides considerably more structure than $\epsilon$-greedy exploration and, as shown in our experiments, substantially improves sample complexity. Second, as we show in this paper, independently trained maximum entropy policies can be composed together by adding their Q-functions, yielding a new policy for the combined reward function that is provably close to the corresponding optimal policy. Composability of controllers, which is not typically possible in standard reinforcement learning, is especially important for real-world applications, where reuse of past experience can greatly improve sample efficiency for tasks that can naturally be decomposed into simpler sub-problems. For instance, a policy for a pick-and-place task can be decomposed into (1) reaching specific x-coordinates, (2) reaching specific y-coordinates, and (3) avoiding certain obstacles. Such decomposable policies can therefore be learned in three stages, each yielding a sub-policy, which can later be combined offline without the need to interact with the environment.

The primary contribution of this paper is a framework, based on the recently introduced soft Q-learning (SQL) algorithm~\cite{haarnoja2017reinforcement}, for learning robotic manipulation skills with expressive neural network policies. We illustrate that this framework provides an efficient mechanism for learning a variety of robotic skills and outperforms state-of-the-art model-free deep reinforcement learning methods in terms of sample efficiency on a real robotic system. Our empirical results indicate that SQL substantially outperforms deep deterministic policy gradient (DDPG) and normalized advantage functions (NAF), which have previously been explored for real world model-free robotic learning with neural networks. We also demonstrate a novel extension of the SQL algorithm that enables composition of previously learned skills
We present a novel theoretical bound on the difference between the policy obtained through composition and the optimal policy for the composed reward function, which applies to SQL and other reinforcement learning methods based on soft optimality. In our experiments, we leverage the compositionality of maximum entropy policies 
in both simulated and physical domains, showing robust learning of diverse skills and outperforming existing state-of-the-art methods in terms of sample efficiency.

\section{RELATED WORK}
\label{sec:related_work}

One of the crucial choices in applying RL for robotic manipulation is the choice of representation. Policies based on dynamic movement primitives and other trajectory-centric representations, such as splines, have been particularly popular, due to the ease of incorporating demonstrations, stability, and relatively low dimensionality~\cite{ijspeert2003learning,ijspeert2013dynamical,peters2006policy,peters2008reinforcement}. However, trajectory-centric policies are limited in their expressive power, particularly when it comes to integrating rich sensory information and reacting continuously to the environment. For this reason, recent works have sought to explore more expressive representations, including deep neural networks, %
at the cost of higher sample complexity.
A number of prior works have sought to address the increased sample complexity by employing model-based methods. For example, guided policy search~\cite{levine2016end} learns a model-based teacher that supervises the training of a deep policy network. 
Other works leverage simulation: Ghadirzadeh \etal~\cite{ghadirzadeh2017deep} considers vision-based manipulation by training perception and behavior networks in simulation and learns only a low-dimensional intermediate layer with real-world interaction. Some works learn vision-based policies completely in simulation and then transfers them into real world \cite{zhang2015towards,sadeghi2016cad,tobin2017domain,andrychowicz2017hindsight}. For this approach to work, the simulator needs to model the system accurately and there needs to be significant variation in the appearance of the synthetic images in order to handle the unavoidable domain-shift between the simulation and the real-world \cite{sadeghi2016cad,tobin2017domain,JamesDJ17,BateuxMLC17}. Another common approach to make reinforcement learning algorithms more sample efficient is to learn from demonstrations~\cite{guenter2007reinforcement,pastor2009learning,theodorou2010reinforcement,vevcerik2017leveraging}, but this comes at the cost of additional instrumentation and human supervision.

Perhaps the most closely related recent work aims to apply deterministic model-free deep RL algorithms to real-world robotic manipulation. Gu et al.~\cite{gu2016continuous} used NAF to learn door opening and reaching by parallelizing across multiple real-world robots, and Ve{\v{c}}er{\'\i}k \etal~\cite{vevcerik2017leveraging} extended DDPG to real-world robotic manipulation by including example demonstrations. 
Our experiments show that SQL can learn real-world manipulation more proficiently and with fewer samples than methods such as DDPG and NAF, without requiring demonstrations, simulated experience, or additional supervision.
We also show that SQL can be extended to enable composition of several previously trained skills. 

Soft Q-learning trains policies that maximize both reward and entropy. Maximum entropy policies have been discussed in many contexts, ranging from applications to inverse reinforcement learning~\cite{ziebart2008maximum}, connections to the Kalman duality and optimal control~\cite{todorov2008general,rawlik2012stochastic,toussaint2009robot}, and to incorporation of prior knowledge for reinforcement learning \cite{fox2015taming}. More recently, several papers have noted the connection between soft Q-learning and policy gradient methods~\cite{haarnoja2017reinforcement,nachum2017bridging,schulman2017equivalence}. While most of the prior works assume a discrete action space, Nachum \etal\cite{nachum2017trust} approximates the maximum entropy distribution with a Gaussian, and, to our knowledge, soft Q-learning~\cite{haarnoja2017reinforcement}
is the only method that approximates the true maximum entropy policy with an expressive inference network, which is essential for tasks requiring diverse behaviour at test time and involving multimodal solutions.

We demonstrate that, not only does SQL lead to more sample-efficient real-world reinforcement learning, but that it can also provide a framework for composing several previously trained policies to initialize compound skills. We derive a novel bound in \autoref{sec:composition} that provides insight into when composition is likely to succeed. Several prior works~\cite{todorov2006linearly,todorov2008general} have studied composition in the context of soft optimality, but typically in a different framework: instead of considering composition where the reward functions of constituent skills are added (i.e., do X \emph{and} Y), they considered settings where a soft maximum (``log-sum-exp'') over the reward functions is used as the reward for the composite skill (i.e., do X \emph{or} Y). We argue that the former is substantially more useful than the latter for robotic skills, since it allows us to decompose a task into multiple objectives that all need to be satisfied.

\section{PRELIMINARIES}
\label{sec:preliminaries}
Soft Q-learning, which we extend in this work for learning composable controllers for real-world robotic manipulation, is based on the framework of maximum entropy reinforcement learning \cite{ziebart2008maximum,kappen2005path,rawlik2012stochastic,todorov2006linearly}. %
In this section, we introduce the formalism of reinforcement learning, describe the soft Q-learning algorithm proposed in prior work~\cite{haarnoja2017reinforcement}, and discuss how it relates to conventional reinforcement learning algorithms, namely DDPG and NAF.

\subsection{Notation}
We will consider an infinite-horizon Markov decision process (MDP), defined by $(\sspace, \aspace, p, \reward)$, where the state space $\sspace$ and the action space $\aspace$ are assumed to be continuous, and the unknown state transition probability $p(\stp|\st,\at)$ represents the probability density of the next state $\stp$ given the current state $\st$ and current action $\at$. The environment emits a bounded reward $\reward(\st,\at)$ on each transition. We will use $\rho_\policy$ to denote the state or state-action marginals of the trajectory distribution induced by a policy $\policy(\at|\st)$. We will drop the time indices from $\st$ and $\at$ and use $(\state, \action)$ and $(\state', \action')$ to denote states and actions for consecutive time steps whenever they can be inferred from the context.

\subsection{Maximum Entropy Reinforcement Learning}
The standard RL objective seeks a policy that maximizes the expected sum of rewards $\sum_t \E{(\st,\at)\sim\rho_\policy}{\reward(\st,\at)}$. In this paper, we consider a more general maximum entropy objective~\cite{ziebart2008maximum,kappen2005path,rawlik2012stochastic,todorov2006linearly}, which favors stochastic policies by augmenting the objective with an expected entropy over $\rho_\policy$:
\begin{align}
\label{eq:maxent_objective}
J(\policy)  = \sum_{t=0}^{T-1} \E{(\st, \at) \sim \rho_\policy}{\reward(\st,\at) \!+\! \alpha\ent(\policy(\sdots|\st))}.
\end{align}
The temperature parameter, $\alpha$, determines the relative importance of the entropy term against the reward, and thus controls the stochasticity of the optimal policy, and %
the conventional objective can be recovered in the limit as $\alpha \rightarrow 0$. The maximum entropy objective has a number of conceptual and practical advantages. First, the policy is incentivized to explore more widely, while giving up on clearly unpromising avenues. Second, the policy can capture multiple modes of optimal behavior. In particular, in problem settings where multiple actions seem equally attractive, the policy will commit equal probability mass to those actions. Lastly, prior work has observed substantially improved exploration from this property~\cite{haarnoja2017reinforcement,schulman2017equivalence}, and in our experiments, we observe that it considerably improves learning speed for real-world robotic manipulation. 
If we wish to extend the objective to infinite horizon problems, it is convenient to also introduce a discount factor $\discount$ to ensure that the sum of expected rewards and entropies is finite. Writing down the precise maximum entropy objective for the infinite horizon discounted case is more involved, and we refer interested readers to prior work (see \eg~\cite{haarnoja2017reinforcement}).

\subsection{Soft Q-Learning}
In this work, we optimize the maximum entropy objective in~\eqref{eq:maxent_objective} using the soft Q-learning algorithm~\cite{haarnoja2017reinforcement}, since, to our knowledge, it is the only existing deep RL algorithm for continuous actions that can capture arbitrarily complex action distributions---a key requirement for policies to be composable, as we discuss in~\autoref{sec:composition}. Soft Q-learning optimizes a Q-function $\Q(\state, \action)$ to predict the expected future return, which include the future entropy values, after taking an action $\action$ at state $\state$ and then following $\policy$. The optimal policy $\policy\opt$ can be expressed in terms of the optimal Q-function as an energy based model (EBM),
\begin{equation}
\label{eq:soft_policy}
\policy\opt(\action|\state) \propto \exp\left(\frac{1}{\alpha}\Q\opt(\state,\action)\right),
\end{equation}
where the Q-function takes the role of negative energy. Unfortunately, we cannot evaluate the action likelihoods, since that would require knowing the partition function, which is intractable in general. However, it is possible to draw samples from this distribution by resorting to a approximate sampling method. To that end, soft Q-learning uses amortized Stein variational descent (SVGD)~\cite{liu2016stein,wang2016learning} to learn a stochastic neural network to approximate samples from the desired EBM~\cite{haarnoja2017reinforcement}. The main benefit of amortizing the cost of approximate sampling into a neural network is that producing samples at test time amounts to a single forward pass, which is fast enough to be done in real-time. 

In addition to learning the optimal policy, we also need to learn the optimal Q-function. It is possible to derive an update rule for the maximum entropy objective in~\eqref{eq:maxent_objective} that resembles the Bellman backup used in conventional Q-learning. This \emph{soft} Bellman operator updates the Q-function according to 
\begin{align}
\label{eq:sql_update}
 \Q(\state, \action) \leftarrow \reward(\state, \action) + \discount \E{\state'\sim p(\state'|\state, \action)}{V(\state')},
\end{align}
where the value function $V(\state)$ is defined as the soft maximum of the Q-function over actions:
\vspace{-6mm}
\begin{align}
\label{eq:soft_value}
\\
V(\state)= \softmax{\action} \Q(\state, \action) = \alpha \log \int_\aspace \exp\left(\frac{1}{\alpha}\Q(\state, \action)\right)d\action\notag.
\end{align}
It is possible to show that the soft Bellman backup is a contraction, and the optimal Q-function is the fixed point of the iteration in~\eqref{eq:sql_update} (see \eg~\cite{haarnoja2017reinforcement}). We can therefore transform any bounded function to the optimal Q-function by iteratively applying the soft Bellman backup until convergence. 
In practice, we represent the Q-function with a parametric function approximator, such as a multilayer neural network, and therefore we cannot perform the soft Bellman backup in its exact from. Instead, we can learn the parameters by minimizing the squared soft Bellman residual, that is, the difference between the left and right hand sides of~\eqref{eq:sql_update}. However, we will make use of the soft Bellman backup in~\autoref{sec:composition}, where we propose a method for composing new policies from existing policies. 

\subsection{Baselines}
In this section, we briefly introduce DDPG and NAF, which are prior model-free reinforcement learning methods that optimize the conventional maximum return objective and have been applied to real-world robotic tasks~\cite{gu2017deeprobot,vevcerik2017leveraging}, and discuss their connection to soft Q-learning.

\subsubsection{Deep Deterministic Policy Gradient}
Deep deterministic policy gradient (DDPG)~\cite{lillicrap2015continuous} is a frequently used prior method that learns a deterministic policy and a Q-function jointly. The policy is trained to maximize the Q-function, whereas the Q-function is trained to represent the expected return of the current policy. The main difference to soft Q-learning is that DDPG replaces the soft maximization in~\eqref{eq:soft_value} with an approximate hard maximum estimated with the current policy. %
Since DDPG uses a deterministic policy, exploration is usually achieved by adding independent noise to the policy output. %
In contrast, SQL explicitly balances exploration and exploitation: the policy can explore more when it is uncertain which actions are good (all Q-values are approximately equal), but will not take actions that are clearly sub-optimal, leading to more informative exploration.

\subsubsection{Normalized Advantage Functions}
Normalized advantage functions (NAF)~\cite{gu2016continuous} consider a Q-function of a special form that is quadratic with respect to the action. The benefit of constraining the form of the Q-function is that it allows closed form computation of the maximizer. Therefore NAF replaces the soft maximum in~\eqref{eq:soft_value} with the exact hard maximum. Similarly, the policy is represented as the deterministic maximum of the Q-function, making it easy to sample actions. In contrast to soft Q-learning, NAF is limited in its representational power and cannot represent diverse or multimodal solutions, which is a major motivation for learning maximum entropy policies.

\section{COMPOSITIONALITY OF MAXIMUM ENTROPY POLICIES}
\label{sec:composition}

Compositionality is a powerful property that naturally emerges from maximum entropy  reinforcement learning. Compositionality means that multiple policies can be combined to create a new policy that simultaneously solves all of the tasks given to the constituent policies. This feature is desirable, as it provides reusability and enables quick initialization of policies for learning complex compound skills out of previously learned building blocks. A related idea has been discussed by Todorov~\cite{todorov2009compositionality}, who considers combining the independent rewards via soft maximization. However, this type of composition corresponds to solving only one of the constituent tasks at a time---a kind of disjunction (e.g., move to the target \emph{or} avoid the obstacle). In contrast, our approach to composition corresponds to a conjunction of the tasks, which is typically more useful (e.g., move to the target \emph{and} avoid the obstacle).

In this section, we will discuss how soft Q-functions for different tasks can be combined additively to solve multiple tasks simultaneously. While simply adding Q-functions does not generally give the Q-function for the combined task, we show that the regret from using the policy obtained by adding the constituent Q-functions together is upper bounded by the difference between the two policies that are being composed. Intuitively, if two composed policies agree on an action, or if they are indifferent towards each other's actions, then the composed policy will be closer to the optimal one.

\subsection{Learning with Multiple Objectives}

Compositionality makes learning far more efficient in multi-objective settings, which naturally emerge in robotic tasks. For example, when training a robot to move objects, one objective may be to move these objects quickly, and another objective may be to avoid collisions with a wall or a person. More generally, assume there are $K$ tasks, defined by reward functions $r_i$ for $i = 1, ..., K$, and that we are interested in solving any subset $\mathcal{C} \subseteq \{1, ..., K\}$ of those tasks simultaneously. A compound task can be expressed in terms of the sum of the individual rewards:
\begin{align}
\label{eq:combined_reward}
\reward_\mathcal{C}(\state, \action)  &= \frac{1}{|\mathcal{C}|}\sum_{i \in \mathcal{C}} \reward_i(\state, \action).
\end{align}
The conventional approach is to solve the compound tasks by directly optimizing this compound reward $r_\mathcal{C}$ for every possible combination $\mathcal{C}$. Unfortunately, there are exponentially many combinations.
Compositionality makes this learning process much faster by instead training an optimal policy $\pi_i\opt$ for each reward $r_i$ and later combining them.

How should we combine the policies $\pi_i\opt$? A simple approach is to approximate the optimal  Q-function of the composed task $\Q_\mathcal{C}\opt$ with the mean of the individual Q-functions:
\begin{align}\label{eq:naive_Q_combination}
Q_\mathcal{C}\opt(\state,\action) \approx \Q_\Sigma(\state, \action)  = \frac{1}{|\mathcal{C}|}\sum_{i \in \mathcal{C}} \Q\opt_i(\state, \action),
\end{align}
where $\Q_\Sigma$ represents an approximation to the true optimal Q-function of the composed task $\Q_\mathcal{C}\opt$.
One can then extract a policy $\policy_\Sigma$ from this approximate Q-function using any policy-extraction algorithm. In conventional reinforcement learning without entropy regularization, we cannot make any guarantees about how close $\Q_\Sigma$ is to $\Q_\mathcal{C}\opt$. However, we show in the next section that, if the constituent policies represent optimal maximum entropy policies, then we can bound the difference between the value of the approximate policy $Q_\mathcal{C}^{\policy_\Sigma}$ and the optimal value $Q_\mathcal{C}\opt$ for the combined task.

\subsection{Bounding the Sub-Optimality of Composed Policies}

To understand what we can expect from the performance of the composed policy $\policy_\Sigma$ that is induced by $\Q_\Sigma$, we analyze how the value of $\policy_\Sigma$ relates to the unknown optimal Q-function $Q_\mathcal{C}\opt$  corresponding to the composed reward $r_\mathcal{C}$. For simplicity, we consider the special case where $\alpha = 1$ and we compose just two optimal policies, given by $\policy_i\opt$, with Q-functions $\Q_i\opt$ and reward functions $\reward_i$. Extending the proof to more than two policies and other values of $\alpha$ is straightforward. We start by introducing a lower bound for the optimal combined Q-function in terms of $Q_i\opt$ and $\policy_i\opt$.
\begin{lemma}
\label{lem:optimal_value_bounds}
Let $Q_1\opt$ and $Q_2\opt$ be the soft Q-function of the optimal policies corresponding to reward functions $r_1$ and $r_2$, and define $Q_\Sigma \triangleq \frac{1}{2}\left(Q_1\opt + Q_2\opt\right)$. Then the optimal soft Q-function of the combined reward $r_\mathcal{C} \triangleq \frac{1}{2}\left(r_1 + r_2\right)$ satisfies
\begin{align}
\resizebox{1.0\hsize}{!}{$
Q_\Sigma(\state, \action) \geq Q_\mathcal{C}\opt(\state, \action) \geq \Q_\Sigma(\state, \action) - C\opt(\state, \action),\ \forall \state \in \sspace, \forall \action \in \aspace,
$}
\end{align}
where $C\opt$ is the fixed point of
\vspace{-6mm}
\begin{align}
\label{eq:bound_constant}
\\
\resizebox{1.0\hsize}{!}{$
C(\state, \action) \leftarrow \discount \E{\state'\sim p(\state'|\state,\action)}{\Dalpha{\policy_1\opt(\voidarg|\state')}{\policy_2\opt(\voidarg|\state')} + \max_{\action'\in\aspace}C(\state', \action')}
$}\notag,
\end{align}
and $\Dalpha{\voidarg}{\voidarg}$ is the R\'enyi divergence of order $\nicefrac{1}{2}$.
\\
\begin{proof}
See~\arefA{app:proof_optimal_value_bounds}.
\end{proof}
\end{lemma}
This lower bound tells us that the simple additive composition of Q-functions never overestimates $\Q_\mathcal{C}\opt$ by more than the divergence of the constituent policies. Interestingly, the constant $C\opt$ can be obtained as the fixed point of the conventional Bellman equation, where the divergence of the constituent policies acts as the ``reward function.'' Thus, $C\opt$ is the ``value'' of an adversarial policy that seeks to maximize the divergence. So far, we bounded $Q_\mathcal{C}\opt$ in terms of the constituent Q-functions, which does not necessarily mean that the composed policy $\policy_\Sigma$ will have a high value. To that end, we use soft policy evaluation~\cite{haarnoja2018soft} to bound the value of the composed policy $Q^{\policy_\Sigma}_\mathcal{C}$:
\begin{theorem}
\label{the:policy_value_bound}
With the definitions in~\autoref{lem:optimal_value_bounds}, the value of $\policy_\Sigma$ satisfies 
\begin{align}
Q^{\policy_\Sigma}_\mathcal{C}(\state, \action) \geq Q\opt_\mathcal{C}(\state, \action) - D\opt(\state, \action),
\end{align}
where $D\opt(\state, \action)$ is the fixed point of
\vspace{-6mm}
\begin{align}
\\
\resizebox{1.0\hsize}{!}{$
D(\state, \action) \leftarrow \discount \E{\state'\sim p(\state'|\state,\action)}{\E{\action'\sim\policy_\Sigma(\action'|\state')}{C\opt(\state',\action') + D(\state',\action')}}
$}\notag.
\end{align}
\vspace{-4mm}\\
\begin{proof}
See~\arefB{app:proof_policy_value_bound}.
\end{proof}
\end{theorem}
Analogously to the discussion of $C\opt$ in~\autoref{lem:optimal_value_bounds}, $D\opt$ can be viewed as the fixed point of a Bellman equation, where now $\discount C\opt(\state, \action)$ has the role of the ``reward function.'' Intuitively, this means that the bound becomes tight if the two policies agree in states that are visited by the composed policy. This result show that we can bound the regret from using the policy formed by composing optimal policies for the constituent rewards. While this bound is likely quite loose, it does indicate that the regret decreases as the divergence between the constituent policies decreases. This has a few interesting implications. First, deterministic policies always have infinite divergence, unless they are identical, which suggests that they are poorly suited for composition. Highly stochastic policies, such as those produced with maximum entropy algorithms like soft Q-learning, have much lower divergences. Intuitively, each policy has a wider distribution, and therefore the policies have more ``overlap,'' thus reducing the regret from composition. As a consequence, we expect maximum entropy RL methods to produce much more composable policies.

\section{EXPERIMENTS}
\label{sec:experiments}

In this section, we show that soft Q-learning can obtain substantially better sample complexity over existing model-free deep RL methods. Our experiments demonstrate fast and reliable learning both in simulated and real-world robotic manipulation domains. We also show how compositionality of maximum entropy policies, discussed in \autoref{sec:composition}, provides a practical tool for composing new compound skills out of previously trained components. We evaluate our method on a pushing task in simulation as well as reaching, Lego block stacking, and combined stacking while avoiding tasks on a real-world Sawyer robot. Videos of all experiments can be found on our website\footnote{\resizebox{0.95\hsize}{!}{\url{sites.google.com/view/composing-real-world-policies/}}} and the code is available on GitHub\footnote{\url{github.com/haarnoja/softqlearning}}.

\subsection{Experimental Setup}
For the simulated tasks, we used the MuJoCo physics engine \cite{todorov2012mujoco}, and for the real-world experiments we used the 7-DoF Sawyer robotic manipulator. The actions are the torque commands at each joint, and the observations are the joint angles and angular velocities, as well as the end-effector position. For the simulated pushing tasks, we also include all the relevant object coordinates, and for the experiments on Sawyer, we include the end-effector position and forces estimated from the actuator currents as observations. We parameterize the Q-function and the policies with a 2-layer neural network with 100 or 200 units in each layer and rectifier linear activations.

\subsection{Composing Policies for Pushing in Simulation}
\label{sec:experiments-compositionality}
Does composing Q-functions from individual constituent policies allow us to quickly learn compound policies? To answer that question, we study the compositionality of policies in a simulated domain, where the task is to move a cylinder to a target location. This simulated evaluation allows us to provide a detailed comparison against prior methods, evaluating both their base performance and their performance under additive composition. We start by learning the Q-functions for the constituent tasks and then combining them by adding together the Q-functions to get $\Q_\Sigma$. To extract a policy from $\Q_\Sigma$, SQL requires training a policy to produce samples from $\exp\left(Q_\Sigma\right)$, and DDPG requires training a network that maximizes $\Q_\Sigma$. In NAF, the optimal action for $\Q_\Sigma$ has a closed-form expression, since the constituent Q-functions are quadratic in the actions. We train the combined DDPG and SQL policies by reusing the data collected during the training of the constituent policies, thus imposing no additional sample cost.

\begin{figure}[tb]
  \centering
  \includegraphics[width=0.75\columnwidth,trim={0 5mm 0 5mm},clip]{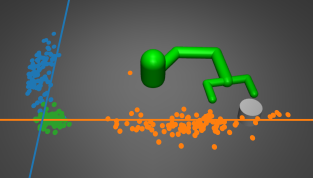}
  \caption{Two independent policies are trained to push the cylinder to the orange line and blue line, respectively. The colored circles show samples of the final location of the cylinder for the respective policies. When the policies are combined, the resulting policy learns to push the cylinder to the lower intersection of the lines (green circle indicates final location). No additional samples from the environment are used to train the combined policy. The combined policy learns to satisfy both original goals, rather than simply averaging the final cylinder location.}
  \label{fig:pusher3dof}
  \vspace{-2mm}
\end{figure}

The constituent tasks in our simulated evaluation require a planar arm to push a cylinder to specific positions along one of the two axes. A policy trained to push the disk to a specific $x$ position can choose any arbitrary $y$ position, and vice-versa. A maximum entropy policy, in this case, would attempt to randomly cover all $y$ positions, while a deterministic one would pick one arbitrarily. The composed policy is formed by combining a policy trained to push a cylinder to a specific $x$ position and a policy trained to push the cylinder to a specific $y$ location, thus solving a combined objective for moving the disk to a particular 2D location on the table. An illustration of this task is depicted in \autoref{fig:pusher3dof}, which shows a compound policy formed by combining a Q-function for pushing the disk the blue line ($x=-1$) and the orange line ($y=-1$). 
The final disk locations for 100 episodes of the final SQL policies are shown with dots of respective colors. The green dots illustrate the disk positions for the combined policy. 
Note that even though the orange policy never moves the disk to any point close to the intersection, the combined policy interpolates to the intended target correctly. %

We trained four policies to push the cylinder to one of the following goals: left ($x = -1$),  middle ($x = 0$), right ($x = 1$), and bottom ($y = -1$) using all three algorithm (SQL, DDPG, NAF). These goals gives three natural compositional objectives: bottom-left ($x=-1, y=-1$), bottom-middle ($x=0, y=-1$), and bottom-right ($x=1, y=-1$). \autoref{tab:pusher_composition} summarizes the error distance of each constituent policy (first four rows), policies trained to optimize the combined objective directly (\eg "push bottom-left"), and composed policies (\eg ``merge bottom-left'').
The policies yielding the lowest error for the individual tasks are emphasized in the table. We see that SQL and DDPG perform well across all combined tasks, whereas NAF is able to combine some policies better (bottom-middle) but fails on others (bottom-right). Note that NAF policies are unimodal Gaussian, which are not well suited for compositions.

\begin{table}[!t]
\renewcommand{\arraystretch}{1.3}
\caption{Simulated Pushing Task: distance from cylinder to goal}
\label{tab:pusher_composition}
\centering
\begin{tabular}{l|rrr}
Task&\multicolumn{1}{c}{SQL}&\multicolumn{1}{c}{DDPG}&\multicolumn{1}{c}{NAF}\\
\hline
push left& 0.13 $\pm$ 0.06& 0.20 $\pm$ 0.01& \bf{0.06} $\pm$ 0.01\\
push middle& 0.07 $\pm$ 0.08& 0.05 $\pm$ 0.03& \bf{0.02} $\pm$ 0.00\\
push right& 0.10 $\pm$ 0.08& 0.10 $\pm$ 0.07& \bf{0.09} $\pm$ 0.06\\
push bottom& 0.08 $\pm$ 0.07& \bf{0.04 $\pm$ 0.02}& 0.17 $\pm$ 0.08\\
\hline
push bottom-left& 0.11 $\pm$ 0.11& 0.24 $\pm$ 0.01& 0.17 $\pm$ 0.01\\
merge bottom-left& \bf 0.11 $\pm$ 0.05& 0.19 $\pm$ 0.10& 0.16 $\pm$ 0.00\\
\hline
push bottom-middle& 0.12 $\pm$ 0.09& 0.14 $\pm$ 0.03& 0.34 $\pm$ 0.08\\
merge bottom-middle& 0.09 $\pm$ 0.09& 0.12 $\pm$ 0.02& \bf 0.02 $\pm$ 0.01\\
\hline
push bottom-right& 0.18 $\pm$ 0.17& 0.14 $\pm$ 0.06& 0.15 $\pm$ 0.06\\
merge bottom-right& 0.15 $\pm$ 0.14& \bf 0.09 $\pm$ 0.10& 0.43 $\pm$ 0.21\\
\end{tabular}
\vspace{-4mm}
\end{table}

\begin{figure}[thpb]
  \centering
  \includegraphics[width=\columnwidth]{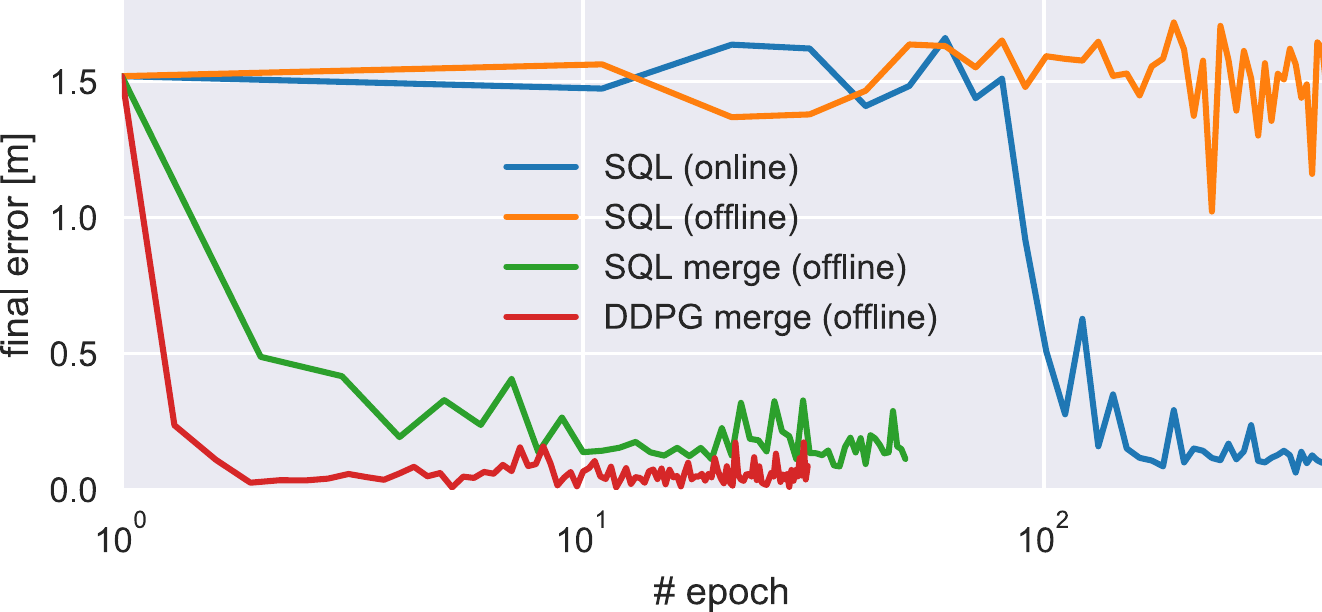}
  \vspace{-6mm}
  \caption{Comparison of the number of iterations needed to extract a policy from a Q-function (red and green) and training a policy using off-policy (orange) or on-policy data (blue). Note that the x-axis is on a log scale. Our proposed method (green and red) extracts a policy from an additively composed Q-function, $\Q_\Sigma$, almost instantly in an offline fashion from off-policy data that were collected for training the constituent Q-functions. In contrast, training a policy from scratch with online SQL (blue) takes orders of magnitude more gradient steps and requires collecting new experience from the environment. Lastly, training a policy on the composed task using offline SQL with pre-collected data fails to converge in a reasonable amount of time (orange).}
  \vspace{-3mm}
  \label{fig:pusher_merge}
\end{figure}

We also compared the different methods of learning a composed task in terms of training time. Example training curves for one of the targets are shown in \autoref{fig:pusher_merge}. Training a policy from a given $\Q_\Sigma$ (red, green) takes virtually no time for any of the algorithms when compared to training a Q-function and policy from scratch (blue). In fact, the combination of NAF policies has a closed form, and it is thus not included in the graph. For comparison, we also trained a Q-function with SQL from scratch on the combined task in an offline fashion (orange), meaning that we only use samples collected by the policies trained on individual tasks. However, offline SQL fails to converge in reasonable time.

Our analysis shows that, both with SQL and DDPG, compound policies can be obtained quickly and efficiently simply by adding together the Q-functions of the individual constituent policies. We can bound the suboptimality of such policies for SQL~\autoref{sec:composition}, but for DDPG our analysis does not directly apply since DDPG is the limiting case $\alpha = 0$. Nonetheless, on simple practical problems like the one in our experiments, even the DDPG Q-functions can be composed reasonably. This suggests that composition is a powerful tool that can allow us to efficient build new skills out of old ones. In the next section, we will see that on more complex real-world manipulation tasks, SQL substantially outperforms DDPG in terms of both learning speed and final performance, and constituent policies trained with SQL can also be reused to form compound skills in the real world.

\subsection{SQL for Real-World Manipulation}

To test the viability of deep RL with maximum entropy policies, we trained a Sawyer robot (\autoref{fig:sawyer}) for reaching different locations and stacking Lego blocks. We also evaluated the compositionality of soft policies by combining the stacking policy with a policy that avoids an obstacle.

\subsubsection{Reaching}
\begin{figure}[tb]
  \centering
  \includegraphics[width=1.\columnwidth]{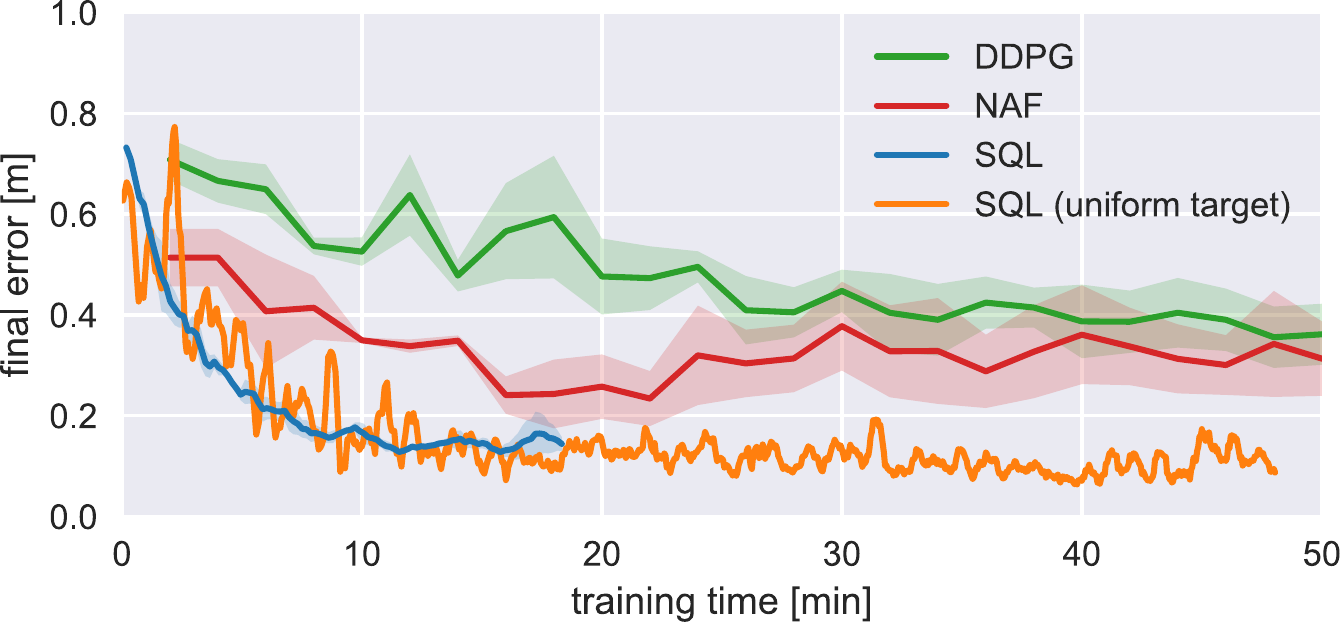}
  \vspace{-6mm}
  \caption{The learning curve of DDPG (green), NAF (red), and SQL (blue) on the Sawyer robot when trained to move its end effector to a specific location. SQL learns much faster than  the other methods. We also train SQL to reach randomly sampled end-effector locations by concatenating the desired location to the observation vector (orange). SQL learns to solve this task just as quickly. SQL curves show the moving average over 10 training episodes.}
  \label{fig:sawyer_reaching_uniform}
\end{figure}

Our first experiment explores how SQL compares with DDPG and NAF in terms of sample complexity. To that end, we trained the robot to move its end-effector to a specified target location in Cartesian space. \autoref{fig:sawyer_reaching_uniform} shows the learning curves for the three methods. Each experiment was repeated three times to analyze the variability of the methods.
Since SQL does not rely on external exploration, we simply show training performance, whereas for DDPG and NAF, we show test performance at regular intervals, with the exploration noise switched off to provide a fair comparison. SQL solves the task in about 10 minutes, whereas DDPG and NAF are substantially slower, and exhibit large variation between training runs.
We also trained a policy with SQL to reach randomly selected points that are provided as input to the policy, with a different target point selected for each episode. Learning this policy was almost as fast as training for a fixed target, but because some targets are easier to reach than others, the policy exhibited larger variation in terms of the final error of the end-effector position.

\subsubsection{Lego Block Stacking}
\begin{figure}[t]
\centering
    \includegraphics[width=0.245\columnwidth]{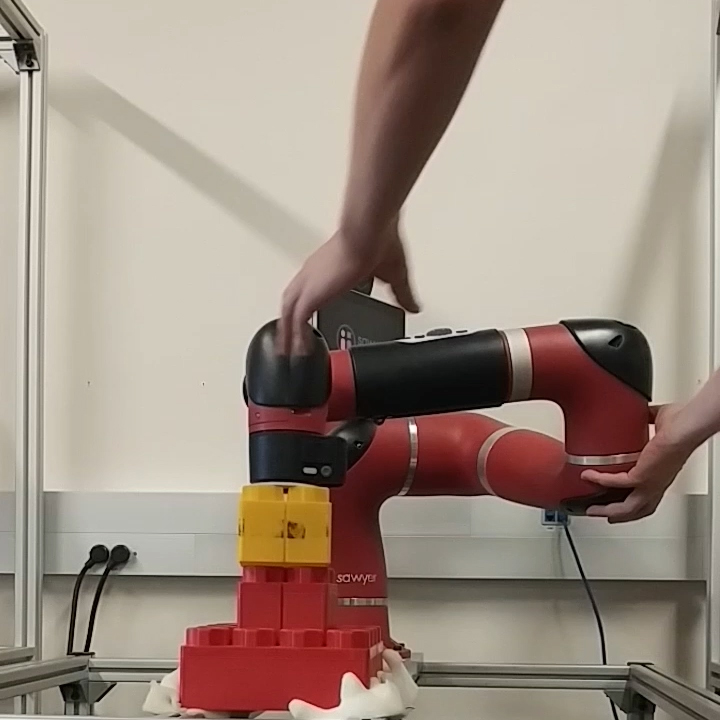}\hfill
    \includegraphics[width=0.245\columnwidth]{figures/sawyer/bullying/bullying03.jpg}\hfill
    \includegraphics[width=0.245\columnwidth]{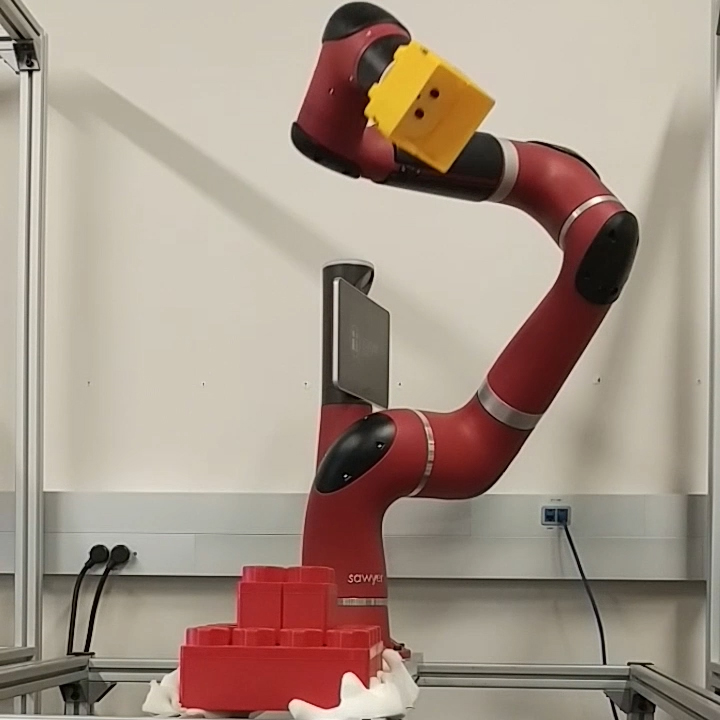}\hfill
    \includegraphics[width=0.245\columnwidth]{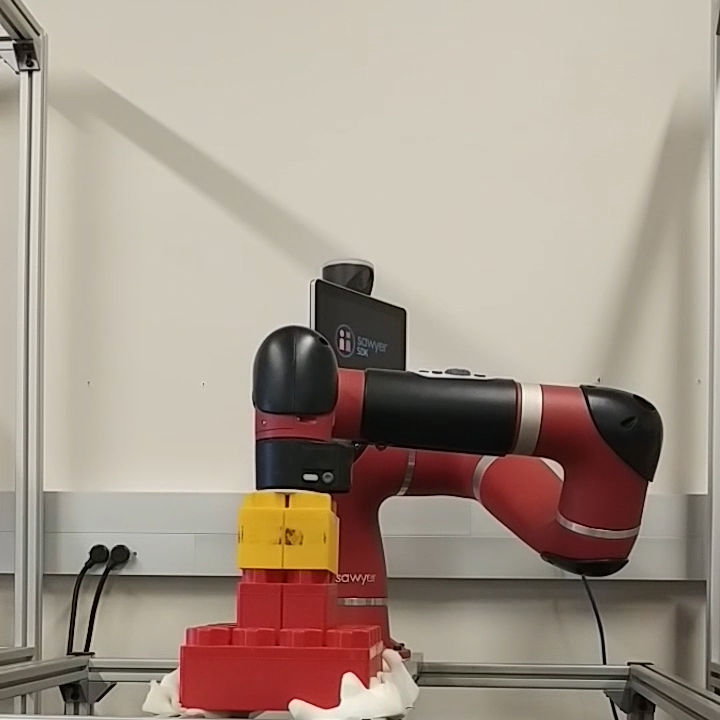}
\caption{A Lego stacking policy can be learned in less than two hours. 
The learned policy is remarkably robust against perturbations: the robot is able to recover and successfully stack the Lego blocks together after being pushed into a state that is substantially different from typical trajectories.}
\label{fig:bullying}
\vspace{-2mm}
\end{figure}

In the next experiment, we used SQL to train a policy for stacking Lego blocks to test its ability to exercise precise control in the presence of contact dynamics (\autoref{fig:sawyer}, left). The goal is to position the end effector, which holds a Lego block, into a position and orientation that successfully stacks it on top of another block. To reduce variance in the end-effectors pose, we augmented the reward function with an additional negative log-distance term that incentivizes higher accuracy when the block is nearly inserted, and we also included a reward for downward force to overcome the friction forces. After half an hour of training, the robot was able to successfully insert the Lego block for the first time, and in two hours, the policy was fully converged. We evaluated the task visually by inspecting if all of the four studs were in contact and observed a success rate of 100\!~\% on 20 trials of the final policy. The resulting policy does not only solve the task consistently, but it is also surprisingly robust to disturbances. To test robustness, we forced the arm into configurations that are far from the states that it encounters during normal execution, and the policy was able to recover every time and eventually solve the task (\autoref{fig:bullying}). We believe the robustness can be attributed to the effective and diverse exploration that is natural to maximum entropy policies, and, to our knowledge, no other prior deep RL work has shown similar results.

\subsubsection{Composing Policies}
\begin{figure}[t] 
\centering
    \includegraphics[width=0.195\columnwidth, trim={0 15mm 0 10mm}, clip]{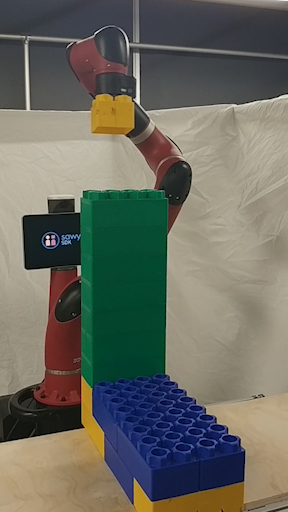}\hfill
    \includegraphics[width=0.195\columnwidth, trim={0 15mm 0 10mm}, clip]{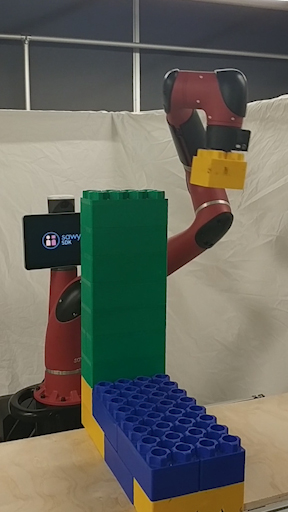}\hfill
    \includegraphics[width=0.195\columnwidth, trim={0 15mm 0 10mm}, clip]{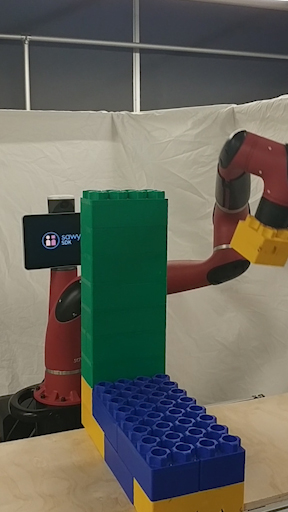}\hfill
    \includegraphics[width=0.195\columnwidth, trim={0 15mm 0 10mm}, clip]{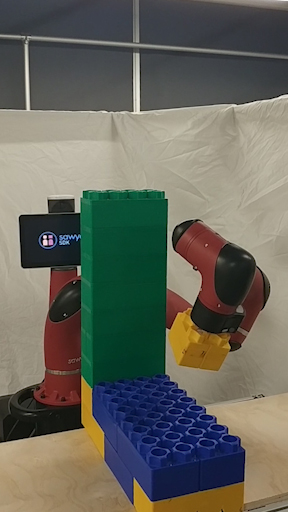}\hfill
    \includegraphics[width=0.195\columnwidth, trim={0 15mm 0 10mm}, clip]{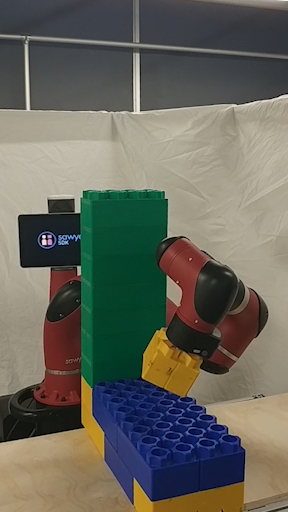}\\
    \vspace{0.5mm}
    \includegraphics[width=0.195\columnwidth, trim={0 15mm 0 10mm}, clip]{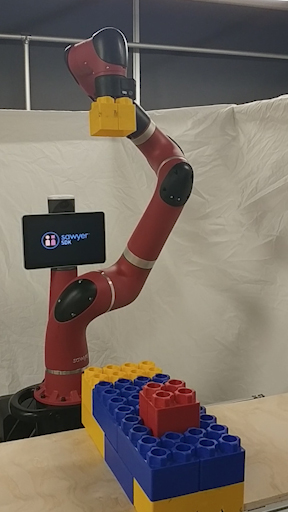}\hfill
    \includegraphics[width=0.195\columnwidth, trim={0 15mm 0 10mm}, clip]{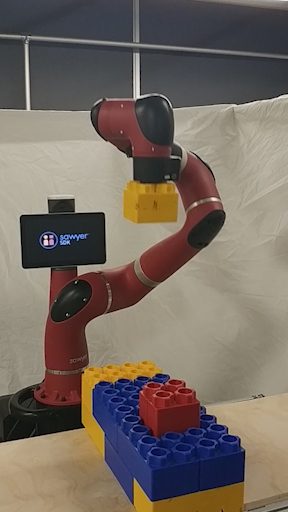}\hfill
    \includegraphics[width=0.195\columnwidth, trim={0 15mm 0 10mm}, clip]{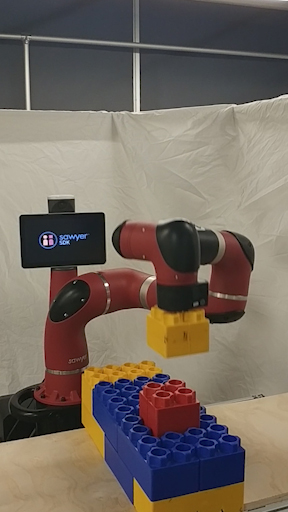}\hfill
    \includegraphics[width=0.195\columnwidth, trim={0 15mm 0 10mm}, clip]{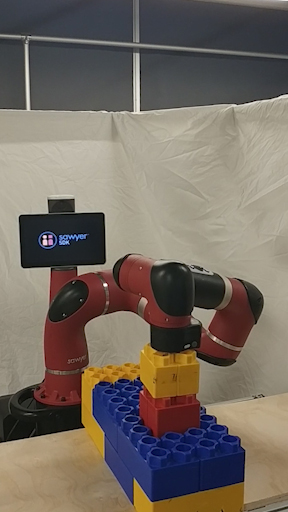}\hfill
    \includegraphics[width=0.195\columnwidth, trim={0 15mm 0 10mm}, clip]{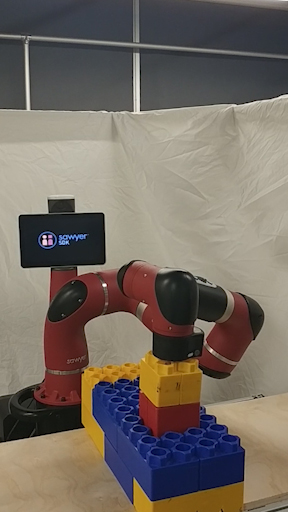}\\
    \vspace{0.5mm}
    \includegraphics[width=0.195\columnwidth, trim={0 15mm 0 10mm}, clip]{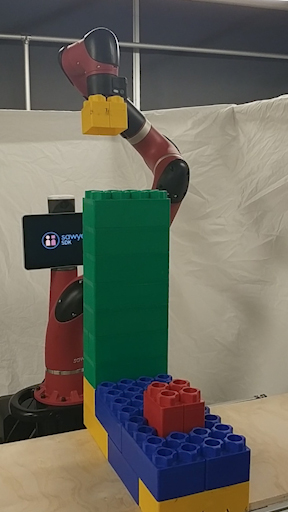}\hfill
    \includegraphics[width=0.195\columnwidth, trim={0 15mm 0 10mm}, clip]{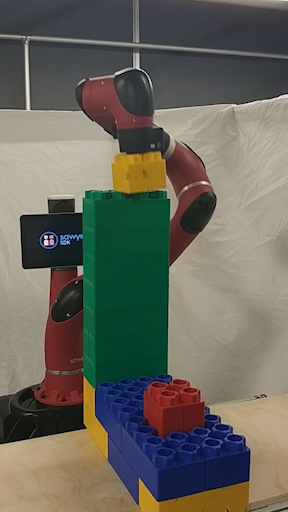}\hfill
    \includegraphics[width=0.195\columnwidth, trim={0 15mm 0 10mm}, clip]{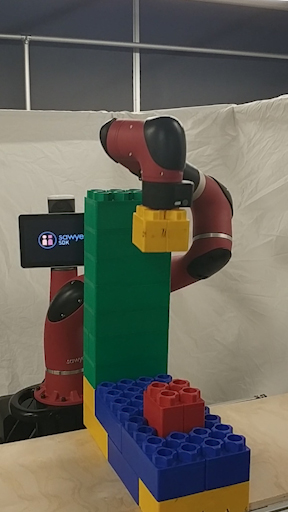}\hfill
    \includegraphics[width=0.195\columnwidth, trim={0 15mm 0 10mm}, clip]{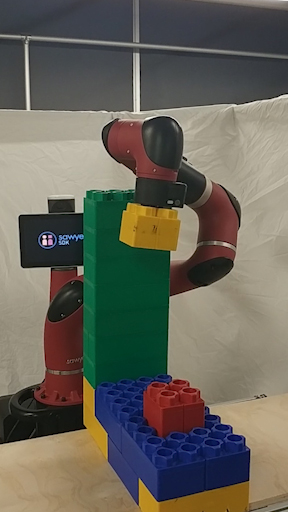}\hfill
    \includegraphics[width=0.195\columnwidth, trim={0 15mm 0 10mm}, clip]{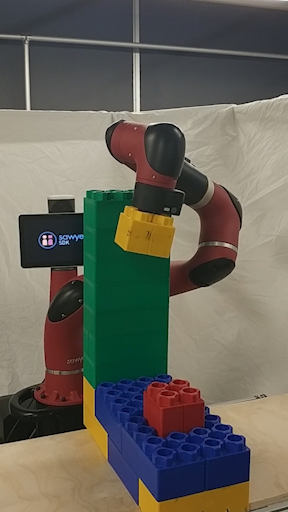}\\
    \vspace{0.5mm}
    \includegraphics[width=0.195\columnwidth, trim={0 15mm 0 10mm}, clip]{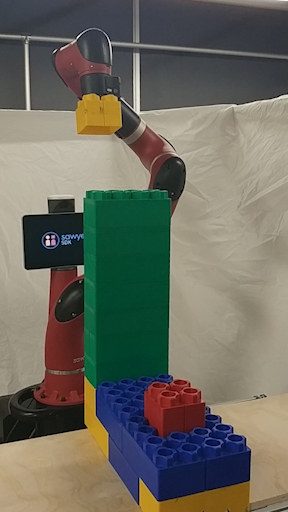}\hfill
    \includegraphics[width=0.195\columnwidth, trim={0 15mm 0 10mm}, clip]{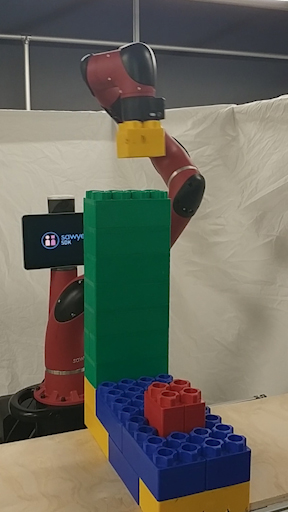}\hfill
    \includegraphics[width=0.195\columnwidth, trim={0 15mm 0 10mm}, clip]{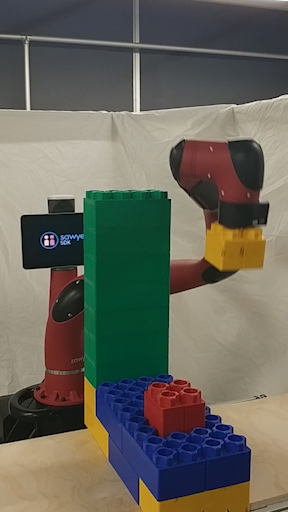}\hfill
    \includegraphics[width=0.195\columnwidth, trim={0 15mm 0 10mm}, clip]{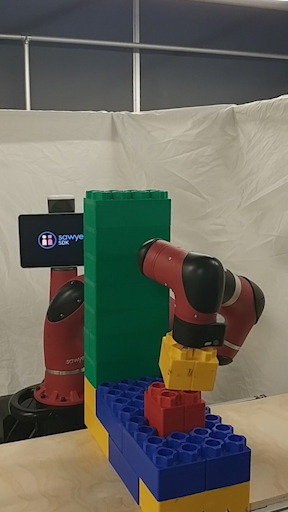}\hfill
    \includegraphics[width=0.195\columnwidth, trim={0 15mm 0 10mm}, clip]{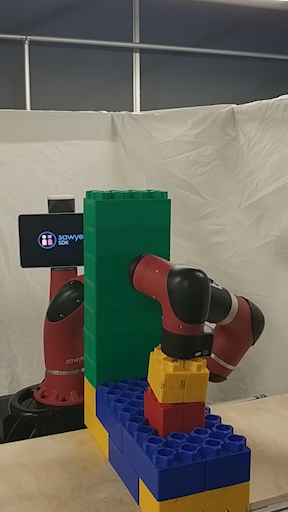}
\caption{To illustrate compositionality on physical hardware, we trained  the Sawyer manipulator to avoid an obstacle (first row) and to stack Lego blocks together (second row). The stacking policy fails if it is executed in the presence of the obstacle (third row). However, by combining the two policies, we get a new combined skills that solves both tasks simultaneously and is able to stack the blocks while avoiding the obstacle. }
\label{fig:sawyer_compositionality}
\vspace{-5mm}
\end{figure}

In the last experiment, we evaluated compositionality on the Sawyer robot. We trained a new policy to avoid a fixed obstacle (\autoref{fig:sawyer_compositionality}, first row) and combined it with the policy to stack Lego blocks (\autoref{fig:sawyer_compositionality}, second row). The avoidance policy was rewarded for avoiding the obstacle and moving towards the target block without the shaping term that is required to successfully stack the two blocks. We then evaluated 1) the avoidance policy, 2) the stacking policy, and 3) the combined policy on the insertion task in the presence of the obstacle by executing each policy 10 times and counting the number of successful insertions. The avoidance policy was able to bring the Lego block close to the target block, but never successfully stacked the two blocks together. The stacking policy collided with the obstacle every time (\autoref{fig:sawyer_compositionality}, third row), but was still able to solve the task 30\!~\% of the time. On the other hand, the combined policy (\autoref{fig:sawyer_compositionality}, bottom row) successfully avoided the obstacle and stacked the blocks 100\!~\% of the time. This experiment illustrates that policy compositionality is an effective tool that can be successfully employed also to real-world tasks.

\section{DISCUSSION AND FUTURE WORK}
\label{sec:conclusions}

In this paper, we discuss how soft Q-learning can be extended to real-world robotic manipulation, both for learning individual manipulation tasks, and for learning constituent tasks that can then be composed into new policies. Our experiments show that soft Q-learning substantially outperforms prior model-free deep reinforcement learning. Soft Q-learning achieves substantially better performance than NAF on a simulated reaching task, including the case where multiple policies are composed to reach to new locations, and outperforms DDPG on a real-world reaching task evaluated on a Sawyer robot. The method exhibits better stability and convergence, and the ability to compose Q-functions obtained via soft Q-learning can make it particularly useful in real-world robotic scenarios, where retraining new policies for each new combination of reward factors is time-consuming and expensive.

In studying the composability of maximum entropy policies, we derive a bound on the error between the composed policy and the optimal policy for the composed reward function. This bound suggests that policies with higher entropy may be easier to compose. An interesting avenue for future work would be to further study the implications of this bound on compositionality. For example, can we derive a correction that can be applied to the composed Q-function to reduce bias? Answering such questions would make it more practical to construct new robotic skills out of previously trained building blocks, making it easier to endow robots with large repertoires of behaviors learned via reinforcement learning.

\noindent \textbf{Acknowledgements.}
We thank Haoran Tang for suggestions on how to improve the proof and Abhishek Gupta for providing the simulated pusher environment. This work was supported by Berkeley DeepDrive.

\bibliographystyle{IEEEtran}
\bibliography{refs}

\onecolumn
\section*{APPENDIX}
\appendices

\subsection{Proof of~\autoref{lem:optimal_value_bounds}}
\label{app:proof_optimal_value_bounds}
\emph{\autoref{lem:optimal_value_bounds}:}
Let $Q_1\opt$ and $Q_2\opt$ be the soft Q-function of the optimal policies corresponding to reward functions $r_1$ and $r_2$, and define $Q_\Sigma \triangleq \frac{1}{2}\left(Q_1\opt + Q_2\opt\right)$. Then the optimal soft Q-function of the combined reward $r_\mathcal{C} \triangleq \frac{1}{2}\left(r_1 + r_2\right)$ satisfies
\begin{align}
\label{eq:q_value_bound_appendix}
Q_\Sigma(\state, \action) \geq Q_\mathcal{C}\opt(\state, \action) \geq \Q_\Sigma(\state, \action) - C\opt(\state, \action),\ \forall \state \in \sspace, \forall \action \in \aspace,
\end{align}
where $C\opt$ is the fixed point of
\vspace{-6mm}
\begin{align}
\\
C(\state, \action) \leftarrow \discount \E{\state'\sim p(\state'|\state,\action)}{\Dalpha{\policy_1\opt(\voidarg|\state')}{\policy_2\opt(\voidarg|\state')} + \max_{\action'\in\aspace}C(\state', \action')}\notag,
\end{align}
and $\Dalpha{\voidarg}{\voidarg}$ is the R\'enyi divergence of order $\nicefrac{1}{2}$.
\\
\begin{proof}
We will first prove the lower bound using induction. Let us define functions $Q^{(0)}(\state, \action) \triangleq Q_\Sigma(\state,\action)$ and $C^{(0)}(\state, \action)\triangleq 0$, and assume $Q^{(k)}\geq Q_\Sigma - C^{(k)}$ for some $k$. Note that this inequality is trivially satisfied when $k=0$. Next, let us apply the soft Bellman backup at step $k$:
\begin{align}
Q^{(k+1)}(\state, \action) &= r_\mathcal{C}(\state, \action) + \discount\E{\state' \sim p(\state'|\state, \action)}{\log\int_\aspace \exp Q^{(k)}(\state', \action')d\action'}\\
&\geq r_\mathcal{C}(\state, \action) + \discount \E{\state'\sim p(\state'|\state, \action)}{\log\int_\aspace \exp \left(\frac{1}{2}\left(Q\opt_1(\state',\action') + Q\opt_2(\state',\action')\right) - C^{(k)}(\state', \action')\right)d\action'}\\
&\geq r_\mathcal{C}(\state, \action) + \discount \E{\state'\sim p(\state'|\state, \action)}{\log\int_\aspace \exp \frac{1}{2}\left(Q\opt_1(\state',\action') + Q\opt_2(\state',\action')\right)d\action' - \max_{\action'}C^{(k)}(\state', \action')}\\
&= r_\mathcal{C}(\state, \action)\!  +\! \discount \E{\state'\sim p(\state'|\state, \action)}{\frac{1}{2}\left(V_1\opt(\state') + V_2\opt(\state')\right) + \log\int_\aspace \sqrt{\policy_1\opt(\action'|\state')\policy_2\opt(\action'|\state')}d\action' - \max_{\action'}C^{(k)}(\state', \action')}\\
&=\frac{1}{2}\left(Q_1\opt(\state, \action) + Q_2\opt(\state, \action)\right) + \discount \E{\state'\sim p(\state'|\state, \action)}{\log\int_\aspace \sqrt{\policy_1\opt(\action'|\state')\policy_2\opt(\action'|\state')}d\action' - \max_{\action'}C^{(k)}(\state', \action')}\\
&= Q_\Sigma(\state, \action) - \discount \E{\state'\sim p(\state'|\state, \action)}{\frac{1}{2}\Dalpha{\policy_1\opt(\voidarg|\state')}{\policy_2\opt(\voidarg|\state')} + \max_{\action'}C^{(k)}(\state', \action')}\\
&= Q_\Sigma(\state, \action) - C^{(k+1)}(\state, \action)
\end{align}
The last form shows that the induction hypothesis is true for $k+1$. Since soft Bellman iteration converges to the optimal soft Q-function from any bounded $Q^{(0)}$, at the limit, we will have 
\begin{align}
Q_\mathcal{C}\opt(\state, \action) \geq Q_\Sigma(\state, \action) - C\opt(\state,\action),
\end{align} 
where the value of $C\opt$ is the fixed point of the following recursion:
\begin{align}
C^{(k+1)} = \discount \E{\state'\sim p(\state'|\state, \action)}{\frac{1}{2}\Dalpha{\policy_1\opt(\voidarg|\state')}{\policy_2\opt(\voidarg|\state')} + \max_{\action'}C^{(k)}(\state', \action')}.
\end{align}
The upper bound follows analogously by assuming $Q^{(k)}\leq Q_\Sigma$, defining $Q^{(0)}=Q_\Sigma$ and noting that the R\'enyi divergence is always positive.
\end{proof}
\vspace{5mm}
\begin{corollary}
\label{cor:optimal_value_bounds}
As a corollary of~\autoref{lem:optimal_value_bounds}, we have
\begin{align}
V_\Sigma(\state) \geq V_\mathcal{C}\opt(\state) \geq V_\Sigma(\state) - \max_{\action}C\opt(\state, \action),\ \forall \state \in \sspace,
\end{align}
where $V_\Sigma(\state)=\log\int_\aspace \exp(Q(\state, \action))d\action$
\\
\begin{proof}
Take the ``log-sum-exp''  of~\eqref{eq:q_value_bound_appendix}.
\end{proof}
\end{corollary}

\subsection{Proof of~\autoref{the:policy_value_bound}}
\label{app:proof_policy_value_bound}

\emph{\autoref{the:policy_value_bound}:}
With the definitions in~\autoref{lem:optimal_value_bounds}, the value of $\policy_\Sigma$ satisfies 
\begin{align}
Q^{\policy_\Sigma}_\mathcal{C}(\state, \action) \geq Q\opt_\mathcal{C}(\state, \action) - D\opt(\state, \action),
\end{align}
where $D\opt(\state, \action)$ is the fixed point of
\vspace{-6mm}
\begin{align}
\\
D(\state, \action) \leftarrow \discount \E{\state'\sim p(\state'|\state,\action)}{\E{\action'\sim\policy_\Sigma(\action'|\state')}{C\opt(\state',\action') + D(\state',\action')}}
\notag.
\end{align}
\vspace{-4mm}\\
\begin{proof}
Let us define functions $Q^{(0)}(\state, \action)\triangleq Q_\mathcal{C}\opt(\state, \action)$ and $D^{(0)}(\state, \action) \triangleq 0$, and assume $\Q^{(k)}(\state, \action) \geq Q_\mathcal{C}\opt(\state,\action) - D^{(k)}(\state, \action)$, which is trivially satisfied for $k=0$. We will use induction to show the inequality holds for any $k$ by applying ``soft policy evaluation'' (see~\cite{haarnoja2017reinforcement}):
\begin{align}
Q^{(k+1)}(\state, \action) &= r_\mathcal{C}(\state, \action) + \discount\E{\state'\sim p(\state'|\state, \action)}{\E{\action'\sim\policy_\Sigma(\action'|\state')}{Q^{(k)}(\state',\action') - \log \policy_\Sigma(\action'|\state')}}\\
&\geq r_\mathcal{C}(\state, \action) + \discount\E{\state'\sim p(\state'|\state, \action)}{\E{\action'\sim\policy_\Sigma(\action'|\state')}{Q\opt_\mathcal{C}(\state',\action') - D^{(k)}(\state', \action') - Q_\Sigma(\state',\action') + V_\Sigma(\state')}}\\
&\geq r_\mathcal{C}(\state, \action) + \discount\E{\state'\sim p(\state'|\state, \action)}{\E{\action'\sim\policy_\Sigma(\action'|\state')}{Q\opt_\mathcal{C}(\state',\action') - D^{(k)}(\state', \action')\! -\! Q\opt_\mathcal{C}(\state',\action')\! -\! C\opt(\state',\action')\! +\! V\opt_\mathcal{C}(\state')}}\\
&\geq r_\mathcal{C}(\state, \action) + \discount\E{\state'\sim p(\state'|\state, \action)}{V_\mathcal{C}\opt(\state') - \E{\action'\sim\policy_\Sigma(\action'|\state')}{C\opt(\state', \action') + D^{(k)}(\state', \action')}}\\
&= Q\opt_\mathcal{C}(\state, \action) - \discount\E{\state'\sim p(\state'|\state, \action)}{\E{\action'\sim\policy_\Sigma(\action'|\state')}{C\opt(\state', \action') + D^{(k)}(\state', \action')}}\\
&= Q\opt_\mathcal{C}(\state, \action) - D^{(k+1)}(\state, \action).
\end{align}
The second inequality follows from~\autoref{lem:optimal_value_bounds} and~\autoref{cor:optimal_value_bounds}. In the limit as $k\rightarrow \infty$, we have $Q_\mathcal{C}^{\policy_\Sigma}(\state, \action) \geq Q_\mathcal{C}\opt(\state, \action) - D\opt(\state, \action)$ (see~\cite{haarnoja2018soft}).
\end{proof}

\end{document}